\documentclass[letterpaper]{article} 
\usepackage{aaai23}  
\usepackage{times}  
\usepackage{helvet}  
\usepackage{courier}  
\usepackage[hyphens]{url}  
\usepackage{graphicx} 
\usepackage{natbib}  
\usepackage{caption} 
\usepackage{algorithm}
\usepackage{algorithmic}

\usepackage{multirow}
\usepackage{booktabs}
\usepackage{amsmath}
\usepackage{subfigure}
\usepackage{newfloat}
\usepackage{listings}
\DeclareCaptionStyle{ruled}{labelfont=normalfont,labelsep=colon,strut=off} 
\floatstyle{ruled}
\newfloat{listing}{tb}{lst}{}
\floatname{listing}{Listing}
\title{AMOM: Adaptive Masking over Masking for \\ Conditional Masked Language Model}
\author{
    Yisheng Xiao\textsuperscript{\rm 1},
    Ruiyang Xu\textsuperscript{\rm 1},
    Lijun Wu\textsuperscript{\rm 2},
    Juntao Li\textsuperscript{\rm 1}\thanks{Corresponding Author},
    Tao Qin\textsuperscript{\rm 2}, 
    Tie-Yan Liu\textsuperscript{\rm 2}, 
    Min Zhang\textsuperscript{\rm 1} 
}
\affiliations{
    \textsuperscript{\rm 1}Institute of Computer Science and Technology, Soochow University \\ 
    \textsuperscript{\rm 2}Microsoft Research Asia

    \{ysxiaoo, ryxu1\}@stu.suda.edu.cn, 
    \{ljt, minzhang\}@suda.edu.cn, 
    \{lijuwu, taoqin, tyliu\}@microsoft.com
}

\usepackage{bibentry}

\begin{document}

\maketitle
\begin{abstract}
Transformer-based autoregressive (AR) methods have achieved appealing performance for varied sequence-to-sequence generation tasks, e.g., neural machine translation, summarization, and code generation, but suffer from low inference efficiency.
To speed up the inference stage, many non-autoregressive (NAR) strategies have been proposed in the past few years.
Among them, the conditional masked language model (CMLM) is one of the most versatile frameworks, as it can support many different sequence generation scenarios and achieve very competitive performance on these tasks.
In this paper, we further introduce a simple yet effective adaptive masking over masking strategy to enhance the refinement capability of the decoder and make the encoder optimization easier.
Experiments on \textbf{3} different tasks (neural machine translation, summarization, and code generation) with \textbf{15} datasets in total confirm that our proposed simple method achieves significant performance improvement over the strong CMLM model.
Surprisingly, our proposed model yields state-of-the-art performance on neural machine translation (\textbf{34.62} BLEU on WMT16 EN$\to$RO, \textbf{34.82} BLEU on WMT16 RO$\to$EN, and \textbf{34.84} BLEU on IWSLT De$\to$En) and even better performance than the \textbf{AR} Transformer on \textbf{7} benchmark datasets with at least \textbf{2.2$\times$} speedup.
Our code is available at GitHub\footnote{\url{ https://github.com/amom-nar/AMOM}}.
\end{abstract}

\section{Introduction}
Transformer-based models~\cite{vaswani2017attention} have been proven effective for various sequence to sequence generation tasks, such as machine translation~\cite{wu2019depth,wu2021r}, text summarization~\cite{savelieva2020abstractive,elsaid2022comprehensive}, dialogue systems~\cite{zhang2020dialogpt,ma2020survey}, code generation~\cite{wang2020towards}, etc. 
Despite the excellent performance of Transformer-based models, they usually adopt the autoregressive (AR) decoding paradigm in which the decoding of a target sequence is decomposed into multi-step predictions in left-to-right order, i.e., the next prediction is conditioned on the previously generated part.
Such an attribute increases the inference time cost linearly with the target sequence length, which is time-consuming for long sequences. 
To alleviate this problem, many recent works explore non-autoregressive (NAR) methods~\cite{gu2018non,qian2020glancing,xiao2022survey} to predict a target sequence in parallel, which can dramatically increase inference speed.
As the cost of increasing decoding speed, NAR models remove the internal dependency of the target sequence and perform each decoding prediction depending entirely upon the source/input sequence. 
Inevitably, the generation quality of NAR methods falls behind their AR counterparts without target-side information in decoding~\cite{gu2018non}.


To achieve a better trade-off between inference speedup and generation quality, the conditional masked language model (CMLM)~\cite{ghazvininejad2019mask} has been proposed and has already become one of the most competitive and widely-used NAR frameworks, which exploits an iterative mask-predict decoding strategy.
In the training stage, CMLM leverages a masked language model objective to generate 
the masked subset of the target sequence in parallel conditioned on the source input and unmasked part in target sequence.
During inference, CMLM first generates the whole target sequence in parallel (the first iteration) and then iteratively masks and predicts low-confidence tokens. 
Based on CMLM, many recent works have achieved performance improvements with advanced enhancement strategies from different perspectives, e.g., improving the inference strategy~\cite{kasai2020parallel,geng2021learning}, benefiting from the AT counterpart~\cite{hao2021multi}, training with better criterion~\cite{ghazvininejad2020aligned,du2021order}, introducing self-correction mechanism~\cite{huang2022improving} and pre-training~\cite{li2022universal}.

In this paper, we further introduce a simple yet very effective strategy to enhance the refinement capability of CMLM without changing the model structure and the inference algorithm, named adaptive masking over masking (AMOM).
Concretely, we present two adaptive masking operations for both the source and target sequence based on the conventional one-time masking in CMLM.
The masking operation for the source sequence can make the encoder optimization easier by adaptively masking a proportion of tokens based on the masked target sequence. 
In contrast, the vanilla CMLM constructs multiple masked target sequences for each source sequence in model training, making the encoder difficult to converge~\cite{guo2020jointly}.
Another potential merit of the source-side masking is to improve the stability of the CMLM model against different decoder inputs by preventing the internal co-adaptation (akin to dropout~\cite{hinton2012improving}).
Moreover, cooperating it with the masking condition of the target sentence can better improve the ability rather than fixed masking. 
Notice that JM-NAT~\cite{guo2020jointly} also explores the source-side masking operation but has a clear difference from our strategy.
It introduces a BERT-like masked language model task on the encoder side to enhance the encoder training, whereas our adaptive strategy does not introduce any extra task and can dynamically capture target-side information.
The target-side adaptive masking operation is presented to enhance the refinement process of CMLM, motivated by the masking ratio changes of the target sequence in different inference iterations, which cannot be captured by the one-time masking. 
Simultaneously, unlike the adaptive target-side masking strategy in GLAT~\cite{qian2020glancing} to achieve curriculum learning, we design the masking strategy specially to encourage the model to perform steadily and conduct refinements effectively. 
We focus on the promotion of each iteration rather than only enhancing the first iteration in GLAT.
More comparisons between our strategy and the counterparts used in GLAT can be found in the experiments part.

Though AMOM is simple, i.e., only two extra masking operations in model training, we find it is surprisingly effective on different sequence generation tasks, including neural machine translation, summarization, and code generation (\textbf{15} datasets in total). 
It achieves state-of-the-art performance on multiple datasets based on the vanilla CMLM, e.g., \textbf{34.62} BLEU score on WMT16 EN$\to$RO, \textbf{34.82} BLEU on WMT16 RO$\to$EN, and \textbf{34.84} BLEU on IWSLT De$\to$En.
AMOM even performs better than the strong autoregressive Transformer on \textbf{7} datasets with at least \textbf{2.2$\times$} speedup.

\section{Methodology}
Our proposed adaptive masking over masking (AMOM) strategy is a simple yet effective add-on for the conditional masked language model (CMLM)~\cite{ghazvininejad2019mask} training, which comprises two adaptive masking operations for the encoder and decoder,
respectively, to enhance the encoder training and the refinement capability of CMLM.
Specifically, we adopt the same encoder-decoder architecture as the CMLM.

\subsection{Conditional Masked Language Model}
\label{sec:cmlm}
A conditional masked language model feeds a source sequence $X$ to the encoder and a target sequence in which part of the tokens are masked by replacing them with the \texttt{[mask]} token to the decoder.
The training objective of CMLM is to learn to predict the masked tokens $Y_{mask}$ in parallel given $X$ and the unmasked tokens $Y_{obs}$ in the rest part of the target sequence, based on the assumption that all target tokens in $Y_{mask}$ are independent of each other, i.e., the prediction of each $Y_{mask}$ token is merely conditioned on $X$ and $Y_{obs}$. 
To eliminate the particularity of $Y_{mask}$, CMLM samples a different number of tokens each time as $Y_{mask}$ from the uniformly distributed number between one to the target length during training, rather than a fixed proportion of the target sequence.
The training objective of CMLM is to maximize:
\begin{equation}
\begin{aligned}
\mathcal{L}_{\text{CMLM}}=\sum_{y_t \in Y_{mask}} \log P(y_t|Y_{obs},X;\theta),
\end{aligned}
\end{equation}
where $\theta$ denotes the trainable parameters of CMLM. 
Unlike AR methods that can automatically decide the decoding end by generating a special \texttt{[EOS]} (end of a sentence) token, typical NAR methods require learning to predict the target length in advance.
CMLM adds a special token \texttt{[LENGTH]} (akin to the \texttt{[cls]} token in BERT) into its encoder to predict the target length. 
During inference, given the input $X$ and the predicted target length, CMLM executes $k$ iterations of mask-predict operation~\cite{ghazvininejad2019mask} to create the final target sequence. At the first iteration, the CMLM predicts the entire $Y$ in parallel fully depending on $X$.
In the next $k-1$ iterations, CMLM repeatedly masks a specific number of low-confidence tokens generated from the last iteration and regenerates them in parallel.

\subsection{Adaptive $X$ Masking}

Basically, CMLM leverages an encoder-decoder structure to achieve sequence to sequence generation, which requires the mutual cooperation between encoder and decoder.
However, during model training, each $X$ will be paired with multiple $Y_{mask}$ due to the uniform masking strategy of CMLM, making the encoder optimization much harder than the decoder. \citeauthor{guo2020jointly} also empirically prove that the convergence speed of the encoder is significant lower than the decoder. Another drawback of conditioning different $Y_{mask}$ on the same $X$ is the internal co-adaptation of $X$, i.e., each prediction of $Y_{mask}$ relies on the whole input sequence, making the decoder less focused on the changes of decoder inputs.

To enhance the encoder training and address the above-mentioned flaws, we propose a simple yet effective adaptive masking for input $X$. Unlike previous research, our proposed adaptive $X$ masking is included in the sequence to sequence generation task, and the number of masked tokens is coordinated with the number of masked $Y$ tokens. 
More concretely, given a training pair $(X, Y)$ in CMLM, where $Y$ will be divided into $Y_{obs}$ and $Y_{mask}$, the masking ratio $\alpha$ of $Y$ can be calculated as $\frac{N_{mask}}{N_{obs}+N_{mask}}$.
$N_{obs}$ and $N_{mask}$ denote the number of tokens in $Y_{obs}$ and $Y_{mask}$, respectively. 
Then, we introduce a mapping function $\varphi(\cdot)$ to decide the masking ratio of $X$ based on the masking ratio in $Y$, i.e., we will randomly mask $\varphi(\alpha) * L_{X}$ tokens in the source sequence, where $L_{X}$ denotes the length of the source sequence. 
Then the training loss of CMLM with adaptive $X$ masking can be computed as:
\begin{equation}
\begin{aligned}
\mathcal{L}_{\text{cmlm}}= -\sum_{y_t \in Y_{mask}} \log P(y_t|Y_{obs},\hat{X};\theta),
\end{aligned}
\label{equ1}
\end{equation}
where $\hat{X}$ refers to the input sequence with $\varphi(\alpha) * L_{X}$ tokens being masked.
We introduce different variations of $\varphi$ in Table~\ref{tab:mask_x} and compare their performance. 


\subsection{Adaptive $Y$ Masking}
\label{sec:aday}
As mentioned above, the superior performance of CMLM-based methods comes from the iterative refinement process, i.e., the previously generated target sequence draft is repeatedly polished by regenerating a specific number of low-confidence tokens in the subsequent iterations.
In seeing the self-correction nature of the refinement process, many recent works introduce a correction objective in CMLM training to enhance its refinement capability e.g., SMART~\cite{ghazvininejad2020semi}, CMLMC~\cite{huang2022improving}.
Unlike these works that introduce extra training objectives and optimize the inference process of CMLM, we present an ultra-simple yet effective adaptive masking operation for $Y$ in model training without any change to the CMLM inference\footnote{More comparisons are given in Appendix.}.
Our strategy is motivated by the quality improvement of predicted tokens along with the refinement iterations, i.e., the proportion of low-confidence tokens (for regeneration in each iteration) from $Y_{mask}$ will gradually decrease along with the refinement iterations, resulting in a varied masking ratio between $Y_{mask}$ and $Y_{obs}$ in the refinement process.

To capture the masking ratio changes in CMLM inference, we add another masking operation (adaptive $Y$ masking) upon the one-time masking in the vanilla CMLM model. 
Specifically, for each training pair $(X, Y)$, $Y$ is divided into $Y_{obs}$ and $Y_{mask}$. CMLM generates the masked tokens based on $Y_{obs}$ and $X$, where the generated result is denoted as $\hat{Y}_{mask}$ to distinguish with $Y_{mask}$.
Then, we compute the correctness ratio of predicted tokens in $\hat{Y}_{mask}$ by comparing with target tokens in $Y_{mask}$, formulated as $\beta = \frac{|\hat{Y}_{mask}=Y_{mask}|}{N_{mask}}$.
Similar to adaptive $X$ masking, we introduce another mapping function $\psi(\cdot)$ to decide the masking proportion of $\hat{Y}_{mask}$ and $Y_{obs}$ tokens. 
Different types of mapping function $\psi(\cdot)$ are experimented in Analysis, and more details are given in Appendix.
We assign a masking probability of $1-\psi(\beta)$ to each token in $\hat{Y}_{mask}$ and a masking probability of $\psi(\beta)$ to each token in $Y_{obs}$.
As a result, the newly masked tokens in the second time denote ${Y'}_{mask}$, and the rest tokens will serve as a new $Y'_{obs}$, for the next iteration. 
The training loss of the new subset ${Y'}_{mask}$ is computed the same as the first-time masking in CMLM, formulated as:
\begin{equation}
\begin{aligned}
\mathcal{L}_{\text{aday}}= -\sum_{y_t \in Y'_{mask}} \log P(y_t|Y'_{obs},\hat{X'};\theta),
\end{aligned}
\label{equ2}
\end{equation}
where $\hat{X'}$ refers to the input sequence with an adaptive masking ratio of $Y'_{mask}$ being masked.

\subsection{AMOM Training and Inference}
\label{sec:inference}
We simply adopt two adaptive masking strategies based on the original CMLM training process.
The training objective of our proposed adaptive masking over masking (AMOM) is the simple combination of $\mathcal{L}_{\text{cmlm}}$ and $\mathcal{L}_{\text{aday}}$ mentioned in Equation~\ref{equ1} and \ref{equ2}, formulated as:
\begin{equation}
\begin{aligned}
\mathcal{L}_{\text{AMOM}}=  \mathcal{L}_{\text{cmlm}} + \mathcal{L}_{\text{aday}},
\end{aligned}
\end{equation}
As for inference, we utilize the same decoding strategy with CMLM.
As mentioned above, we utilize a special token \texttt{[LENGTH]} in the encoder to predict the target length in advance.
Inevitably, there is a deviation between the predicted length and the ground-truth length.
Thus, we also consider selecting the translation with the highest probability with different target lengths to obtain better results.
Given the target length $L_Y$ and the total number of refinement iterations $T$, the model performs generation based on the fully masked decoder input (i.e., empty $Y_{obs}$) at the first iteration.
In the next $T-1$ iterations, a specific number of low-confidence tokens will be masked and re-generated. 
The number of masked tokens in each iteration can be computed as $n=\frac{T-t}{T}*L_Y$, where $t$ denotes the current iteration number.
Given the number of masked tokens, the model will select them based on the output probability of each token, where tokens with the lowest probability will be masked, and their scores will be updated in the next iteration. 
   
\label{sec:decoding}
\section{Experiments}
\label{sec:exp}

\begin{table*}[!]

\centering
\small
    \begin{tabular}{llccccccc}
    \toprule
   \multicolumn{2}{l}{\multirow{2}{*}{\textbf{Model}}} &  \multirow{2}{*}{\textbf{Iterations}} & \multicolumn{2}{c}{\textbf{WMT16}} & \multicolumn{2}{c}{\textbf{WMT14}} & \multirow{2}{*}{\textbf{Speedup}} \\
  &  & & \textbf{EN$\to$RO} & \textbf{RO$\to$EN} & \textbf{EN$\to$DE} & \textbf{DE$\to$EN} & \\
    \midrule
    \multicolumn{2}{l}{\textbf{AR} Transformer~\cite{vaswani2017attention}*}
    & $N$ & 34.23 & 34.28 & 28.41 & 32.28 &1.0x \\
\midrule
    \multirow{6}{*}{\rotatebox{90}{\textbf{Full NAT}}}&
    NAT-FT~\cite{gu2018non} & 1 & 27.29 & 29.06 & 17.69 & 21.47 & 15.6×\\
    & AXE~\cite{ghazvininejad2020aligned}
    & 1 & 31.54 & 30.75  & 23.53 & - &15.3x\\
    &OAXE~\cite{du2021order}
    & 1 & 33.3 & 32.4  & 26.1 & - &15.3x\\
    &GLAT~\cite{qian2020glancing}
    & 1 & 32.87 & 33.51  & 26.55 & 31.02&15.3x\\
    &FullyNAT~\cite{gu2020fully}
    & 1& 33.71 & 34.16  & 27.20 & 31.39& 16.8x\\
    &DSLP~\cite{huang2022non} 
    & 1 & 34.17 & 34.60 & 27.02 & 31.61 & 14.8x \\
    &DAT~\cite{huang2022directed} 
    & 1 & - & - & 27.49 & 31.37 & 13.9x\\
\midrule
    \multirow{3}{*}{\rotatebox{90}{\textbf{Iterative}}} &
    Refine-NAT~\cite{lee2018deterministic}
    & 10 & 27.11 & 30.19 & 21.61 & 25.48 &1.5x \\
    &LevenshteinNAR~\cite{gu2019levenshtein} 
    & \textgreater 7 & 33.02 & - & 27.73 & - &4.0x \\
    &DisCo~\cite{kasai2020parallel}
    & 3.1 & 33.25 & 33.22 & 27.34 & - &3.5x \\
\midrule
  \multirow{8}{*}{\rotatebox{90}{\textbf{CMLM-Based}}} 
    &CMLM~\cite{ghazvininejad2019mask}*
    & 10 & 33.46 & 33.83 & 27.21 & 31.03 &2.3x\\
    &SMART~\cite{ghazvininejad2020semi}
    & 10 & 33.85 & 33.53 & 27.65 & 31.27 &1.7x\\
    &JM-NAT~\cite{guo2020jointly} 
    & 10 &33.52&33.72&27.69 &\textbf{32.24}&-\\
    &RDP~\cite{ding2020understanding}
    & 10 & 33.7 & - & 27.8 & - &1.5x\\
    &LFR~\cite{ding2021rejuvenating}
    & 10 & - & 33.9 & 27.8 & - &1.5x\\
    &MvSR-NAT~\cite{xie2021mvsr} 
    & 10 & 33.38 & 33.56 & 27.39 & 31.18 &3.8x    \\
    &CORR~\cite{huang2022improving}
    & 10 & 34.31 & 34.08 & 28.19 & 31.31 &-  \\
    &CMLMC~\cite{huang2022improving}
    & 10 & 34.57 & 34.13  & \textbf{28.37} & 31.41 &-\\
    \midrule
    \multicolumn{2}{l}{\textbf{Ours} AMOM}  
   & 10 & \textbf{34.62} & \textbf{34.82} & 27.57 &31.67 &2.3x  \\
    \bottomrule
    \end{tabular}
\caption{Results on 4 WMT machine translation tasks. ``*'' denotes the results of our implementations.  }
\label{tab:wmtresults}
\end{table*}

\begin{table*}[ht]
\centering
\small
\begin{tabular}{l | c c c c | c c}
\toprule
\textbf{Model} & \textbf{En$\leftrightarrow$De}  & \textbf{En$\leftrightarrow$Fr}  & \textbf{En$\leftrightarrow$Zh}  & \textbf{En$\leftrightarrow$Es}&
\textbf{Avg} & \textbf{Speedup}\\
\midrule
Transformer & 28.71/34.68  & 36.2/37.0 & 25.7/18.2 & 37.8/39.5 & 32.22 &1.0x\\
\midrule
CMLM &27.77/33.87&35.2/35.0 &26.0/17.9 &37.1/39.0&31.48 &2.2x\\
\midrule
AMOM  &28.41/\textbf{34.84} &35.6/36.3 &\textbf{26.1}/\textbf{18.4} &\textbf{38.0}/\textbf{39.8}&32.18 &2.2x\\
\bottomrule
\end{tabular}
\caption{
Results on $8$ IWSLT datasets. 
Numbers before and after ``/'' denote BLEU scores from and to English directions.
}
\label{tab:iwslt_all}
\end{table*}

To evaluate our AMOM method and show its universal impact on various sequence generation tasks, we conduct experiments on natural machine translation, summarization, and code generation tasks.
\subsection{Datasets}
For machine translation, we conduct experiments both on IWSLT and WMT datasets, which are widely used for NMT tasks. The datasets from IWSLT competitions contain 4 language pairs (170k pairs), see details in Table~\ref{tab:iwslt_all}. For WMT datasets, we choose two language pairs which are widely used in non-autoregressive machine translation task, WMT16 English$\rightarrow$Roman (0.6M pairs) and WMT14 English$\rightarrow$German (4.5M pairs) tasks.
Following previous works on non-autoregressive machine translation, we apply sequence-level knowledge distillation~\cite{kim2016sequence,zhou2019understanding} for all datasets. For WMT datasets, we use the same distilled data as the same as CMLM~\cite{ghazvininejad2019mask}. Then, we amalgamate the raw and distilled data as our final training data, following~\cite{ding2020understanding}. For all IWSLT datasets, we train the teacher model with Transformer$_{small}$, and use the generated results as the distilled data. Then, we train our AMOM on distilled data. 
For summarization task, we use the XSUM dataset~\cite{narayan2018don} which contains 204,045/11,332/11,334 online articles and single sentence summary pairs from the British Broadcasting Corporation for training/validation/test. We preprocess the dataset, following~\cite{lewis2020bart}.
For code generation task, we use Py150 dataset~\cite{raychev2016probabilistic} and use GitHub-Java dataset~\cite{allamanis2013mining}. We use the Python official library tokenizer\footnote{\url{https://docs.python.org/3/library/tokenize.html}} and Javalang\footnote{\url{https://github.com/c2nes/javalang}} to split the datasets into lines of codes. Then we use a sliding context window to adopt 10-lines of code tokens as the source sentences and the next 4-lines as the target sentences. We follow~\cite{wang2020towards} to process the dataset to transform some special tokens as \texttt{[str]} token (without bpe).

\subsection{Settings}
All experiments are done using the Fairseq library~\cite{ott2019fairseq}. 
Following previous settings~\cite{ghazvininejad2019mask}, we use the standard Transformer$_{base}$ configuration on WMT datasets and standard Transformer$_{small}$ configuration on IWSLT datasets for both auto-regressive and non-autoregressive experiments. During AMOM training, we follow the hyper-parameters in CMLMC~\cite{huang2022improving} for WMT14 En$\leftrightarrow$De and follow the hyper-parameters of CMLM realization in Fairseq\footnote{\url{https://github.com/facebookresearch/fairseq/tree/main/examples/nonautoregressive_translation}} for the other datasets.
During inference, we average the 5 best checkpoints chosen by validation BLEU scores as our final model and set the length beam as 3/5 for IWSLT/WMT datasets.
For XSUM, we choose Transformer$_{base}$ with embedding dimension 768 and follow the training schedule applied in NMT. During our training, we make a specific modification of the hyper-parameters referring to~\cite{lewis2020bart}.
During inference we follow the process in~\cite{qi2021bang}, where the same consecutive tokens will be merged to avoid repeated n-gram tokens. 
For code generation tasks, we choose Transformer$_{base}$ with embedding size 512 and follow the original training schedule. We make a specific modification of the hyper-parameters referring to~\cite{liu2022non}. 
For all datasets, we set the limits ratio of adaptive $X$ from 10\%-30\% and adaptive $Y$ from 20\%-80\%, and select a linear mapping function to decide the masking ratios.
More details about training are presented in Appendix.

\subsection{Main Results} 
\noindent{\textbf{Natural Machine Translation.}} 
Following previous works, we evaluate the performance with BLEU~\cite{papineni2002bleu} for WMT datasets and IWSLT En$\leftrightarrow$De dataset, and for the other IWSLT datasets, we use SacreBLEU~\footnote{\url{https://github.com/mjpost/sacrebleu}}~\cite{post2018call,wu2021r}. Speedup is measured by $L_{1}^{\text{GPU}}$ following the previous work~\cite{kasai2020deep,gu2020fully,helcl2022non}. Table \ref{tab:iwslt_all} presents the results on 8 IWSLT datasets, we compare our AMOM with original CMLM and strong Transformer (AR) baseline. First, a significant improvement can be found over the original CMLM on all datasets, with about 0.7 BLEU on average. More excitingly, compared with the strong Transformer (AR) baseline, our AMOM has achieved better performance on five datasets, and only a tiny gap (0.04 BLEU) still exists on average. We show our results in Table~\ref{tab:wmtresults}
for WMT datasets, we compare our approach with various iterative NAR models, including two popular fully NAR models.
We re-run the experiments of CMLM with the same settings in AMOM to avoid inconsistency. After applying our simple yet effective methods to the traditional CMLM framework, we achieved state-of-the-art (SOTA) BLEU score on WMT16 En$\rightarrow$Ro (34.62) and Ro$\rightarrow$En (34.82) with 10 iterations. For the WMT14 En$\leftrightarrow$De dataset, AMOM also outperforms most of the baselines on De$\rightarrow$En (31.67). On the En$\rightarrow$De dataset, AMOM only gains 0.36 BLEU improvement compared with CMLM and a comparable score compared with strong CMLM-Based baselines. 
This might be because our adaptive $X$ strategy hurts the performance in the first iteration to some extent. Note that AMOM is complementary to other effective tricks applied in CMLM, and stronger results can be expected by combining our adaptive masking strategies with their methods. 

\begin{table}[!]
\centering
\small
\begin{tabular}{l c c c }
\toprule
\textbf{Model} & \textbf{ROUGE-1}& \textbf{ROUGE-2} &\textbf{ROUGE-L}\\
\midrule
Transformer & 30.66 & 10.80 & 24.48 \\
\midrule
\textbf{Without pretrain} \\
vanilla NAT	& 24.04 & 3.88	&20.32 \\
InsertNAR &	17.65&	5.18&	16.05 \\
Levenshitein	&25.33	&7.40	&21.48 \\
Disco	&26.85	&6.86	&21.72 \\
POSPD	&27.39	&7.26	&22.15 \\
CMLM*	&25.80	&6.31	&20.45 \\
AMOM*	&\textbf{31.59}	&\textbf{9.30}	&\textbf{24.98} \\
\midrule
\textbf{With pretrain} \\
BANG	&34.71	&11.71	&29.16 \\
MIST	&34.63	&11.29	&28.70 \\
ELMER & \textbf{37.30} & \textbf{13.17}  & \textbf{29.92} \\
\bottomrule
\end{tabular}
\caption{
Results on XSUM for the text summarization task. ``*'' denotes the results of our implementations.
}
\label{tab:xsummore}
\end{table}

\begin{table}[!]
\centering
\small
\begin{tabular}{lcccccc}
\toprule
\multirow{2}{*}{\textbf{Model}} & \multicolumn{3}{c}{\textbf{Python}} & \multicolumn{3}{c}{\textbf{JAVA}} \\
& \textbf{Iter.} & \textbf{BLEU} & \textbf{ES} & \textbf{Iter.} & \textbf{BLEU} & \textbf{ES}\\
\midrule
\multirow{2}{*}{CMLM} 
& 4 & 49.61 & 69.58 & 4 & 60.54 &76.68\\
& 10 & 53.44 &70.42 & 10 & 62.82 & 77.24\\
\multirow{2}{*}{AMOM} 
& 4 & 50.57 & 70.22 & 4 & 62.86 & 76.61 \\
& 10 & \textbf{56.50} &\textbf{71.38} & 10 & \textbf{65.43} & 77.17\\
\bottomrule
\end{tabular}
\caption{Results on Py150 and Github-Java dataset. }
\label{tab:code_gen}
\end{table}

\noindent{\bf Summarization.} 
See Table~\ref{tab:xsummore}, the performance is evaluated by ROUGE F1 score~\cite{lin2002manual}.
Specifically, we report the unigram ROUGE-1 and bigram ROUGE-2 overlap to assess the
informativeness, and the longest common subsequence ROUGE-L score to assess the fluency.
We compare our AMOM with the original CMLM and several NAR baseline models,
including vanilla NAT~\cite{gu2018non}, InsertNAR~\cite{stern2019insertion}, Levenshitein~\cite{gu2019levenshtein}, Disco~\cite{kasai2020parallel}, POSPD~\cite{yang2021pos}, CMLM~\cite{ghazvininejad2019mask}, BANG~\cite{qi2021bang}, MIST~\cite{jiang2021improving}, ELMER~\cite{li2022elmer}.
Results show that AMOM outperforms all other NAR models without pre-training. 
Since pre-training always benefits summarization task a lot, models with pre-training
achieve significant performance improvements. 
Notice that AMOM can also be applied to the pre-training and finetune stage, we believe it also works to improve the performance.

\noindent{\bf Code Generation.} The performance is evaluated by BLEU and ES~\cite{wang2020towards}, which measure character-level edit similarity and $n$-gram level precision between the target codes and generated codes, respectively. We also report the results of different iterations in Table~\ref{tab:code_gen}. Our AMOM outperforms the original CMLM with different iterations and gains better improvements during refinements.
\subsection{Analysis} 
\label{sec:x_mapping}

\noindent{\textbf{The Mapping Function of Adaptive $X$ Masking.}} In this subsection, we exhibit exhaustive experiments to explore encoder masking strategies and how to affect the model performance. 
In particular, we analyse the effects of different mapping functions,
these strategies can utilize decoder masking ratio $\alpha_{dec}$ to obtain encoder masking ratio $\alpha_{enc}$:
\begin{itemize}
\setlength{\itemsep}{0pt}
 \item $\varphi_{linear}$: $\alpha_{enc} = (b-a)\alpha_{dec} + a$;

 \item $\varphi_{convex}$: $\alpha_{enc} = (b-a)\alpha_{dec}^2+b$;

 \item $\varphi_{concvae}$: $\alpha_{enc}= (a-b)\alpha_{dec}^2+2(b-a)\alpha_{dec}+b $;

\item $\varphi_{ladder}$: $\alpha_{enc} = 
    a- \lceil\frac{\alpha_{dec}}{a-b+0.1}\rceil $,
\end{itemize}
where $a$ and $b$ are two hyper-parameters controlling the masking limits, and the specific curves corresponding to the above mapping function are presented in Appendix. 
The results are shown in Table~\ref{tab:mask_x}, and it is worth noting that the above experiments are based on the CMLM model and IWSLT14 De$\to$En dataset for clear contrast.
Early experiments show that encoder masking can boost the model performance, and at $\alpha_{enc}=0.2$, the encoder masked model performs best when using the fixed masking strategy, results are shown in Appendix. That is why we design the mapping function to limit the masking raio around $0.2$. Firstly, we take linear mapping functions as our priority. Fortunately, linear mapping has been proved by comprehensive experiments that it is indeed one of the most effective implementations to boost the performance.
Besides, the results are consistent with our intuition that the more tokens in $Y$ are masked, the few tokens in $X$ should be masked to keep the masking ratio balanced.
We also have briefly tried a few alternative implementations beyond linear mappings, but without achieving further performance improvement.




\noindent{\textbf{The Effect of Adaptive $X$ Masking.}}
We also compare our adaptive X masking strategy with several related works to further show its effectiveness.
Since JM-NAT~\cite{guo2020jointly} also introduces masking operation in $X$, we also conduct experiments to compare AMOM and their bert-like masking. Also, they introduce an auxiliary MLM training objective to improve the encoder, we further verify if this can combine with AMOM, 
see Table \ref{tab:jmnat}. Notice that we keep the decoder-side the same as vanilla CMLM (without adaptive $Y$ masking in AMOM and $n$-gram loss in JM-NAT) to make a fair comparison of encoder-side. Results show that this MLM training objective can also improve AMOM slightly, but seems less related to our assumption and purpose. Besides, we can find adaptive $X$ outperforms the bert-like masking for CMLM.
Also, we find that the adaptive $X$ masking operation is similar to a data augmentation strategy (such as cutoff~\cite{shen2020simple}), and specially designed to improve the refinements ability of CMLM.  To better analyze them, we also compare adaptive $X$ masking with several common data augmentation strategies (including cutoff). 
Since fixed masking is similar to token cutoff, we conduct experiments with span cutoff and mix cutoff. We also compare with some other strategies (such as random delete, random replace). Results show that adaptive $X$ masking outperforms all other operations on $X$, while various traditional strategies can boost vanilla CMLM to some extent.

\begin{table}[!]
\centering
\small
\centering
    \begin{tabular}{l|c|c}
    \toprule
    \textbf{Strategy} & $\alpha_{enc} $ & \textbf{BLEU} \\
    \midrule
    \multirow{6}{*}{Linear} 
    & $\varphi_{linear}(\alpha_{dec}, 0.25, 0.15)$ &  34.20 \\
    & $\varphi_{linear}(\alpha_{dec}, 0.3, 0.1)$ & \textbf{34.48}  \\
    & $\varphi_{linear}(\alpha_{dec}, 0.35, 0.15)$ & 34.30  \\
    & $\varphi_{linear}(\alpha_{dec}, 0.4, 0.1)$ & 34.40  \\
    & $\varphi_{linear}(\alpha_{dec}, 0.1, 0.3)$ & 33.64  \\
    & $\varphi_{linear}(\alpha_{dec}, 0.1, 0.4)$& 33.76\\
    \midrule
    Convex & $\psi_{convex}(\alpha_{dec}, 0.3, 0.1)$  & 33.55 \\ \midrule
    Concave & $\psi_{concave}(\alpha_{dec}, 0.3, 0.1)$  & 33.96 \\ \midrule
    Ladder & $\psi_{ladder}(\alpha_{dec}, 0.3, 0.1)$  & 34.17 \\
    \bottomrule
    \end{tabular}
\caption{The BLEU scores of adaptive $X$ masking strategy.}
\label{tab:mask_x}
\end{table}

\begin{table}[!]
\centering
\small
\begin{tabular}{l c|l c}
\toprule
Method & BLEU & Method & BLEU\\
\midrule
CMLM & 33.87 & CMLM & 33.87\\
+ adax & \textbf{34.48} &  + mix cutoff &33.96\\
+ adax+mlmloss & \textbf{34.57} &  + span cutoff &33.93\\
+ jm-nat & 34.13 & + random replace & 34.13\\
+ jm-nat+mlmloss & 34.21 & + random delete & 33.95\\
\bottomrule
\end{tabular}
\caption{
Comparison between adaptive $X$ masking and related methods.
}
\label{tab:jmnat}
\end{table}

\noindent{\textbf{The Mapping Function of Adaptive $Y$ Masking.}}
\label{sec:aday_exp}
We also experiment with different masking strategies when applied to the decoder side in a two-step training scheme.
We try same adaptive mapping function and denoted as $\psi_{linear}$, $\psi_{convex}$, $\psi_{concvae}$, and $\psi_{ladder}$ to obtain masking ratio $\alpha_{dec}$.
Specifically, we can calculate $\alpha_{dec}$ based on randomly sampled variable $\beta$ which is correctness ratio predicted by first step training as mentioned above : $\alpha_{dec} = \psi_{linear}(\beta, a, b) = (b - a)\beta + a$.
Unlike the encoder masking mapping function, we choose a large masking ratio range because there exist various conditions of masking ratios and tokens confidence during inference. 
The schedule curves are also shown in Appendix.
Table~\ref{tab:mask_y_relation} lists the results of several adaptive decoder masking strategies. Notice that we achieve all results here with a linear mapping $\varphi_{linear}(\alpha_{dec}, 0.3, 0.1)$ for source-side masking.
The simple linear mapping function achieves the best performance, and 
the large masking ratio range seems better.
Besides, a high correctness ratio always indicates high token confidence, and then fewer tokens in $\hat{Y}_{mask}$ will be masked in the next iteration. Our adaptive $Y$ masking strategy matches the inference strategy of the original CMLM.


\begin{table}[!]
\centering
\small
\centering
    \begin{tabular}{l|c|c}
    \toprule
    \textbf{Strategy} & $\alpha_{dec} $ & \textbf{BLEU} \\
    \midrule
    \multirow{6}{*}{Linear} 
    & $\psi_{linear}(\beta, 0.1, 0.9)$ & 34.65 \\
    & $\psi_{linear}(\beta, 0.2, 0.8)$ & \textbf{34.84}  \\
    & $\psi_{linear}(\beta, 0.3, 0.7)$ & 34.79 \\
    & $\psi_{linear}(\beta, 0.2, 0.5)$ & 34.62 \\
    & $\psi_{linear}(\beta, 0.5, 0.8)$ & 34.77\\
    & $\psi_{linear}(\beta, 0.8, 0.2)$ & 34.61 \\
    \midrule
    Convex & $\psi_{convex}(\beta, 0.2, 0.8)$  & 34.80 \\ \midrule
    Concave & $\psi_{concave}(\beta, 0.2, 0.8)$  & 34.59 \\ \midrule
    Ladder & $\psi_{ladder}(\beta, 0.2, 0.8)$  & 34.75 \\
    \bottomrule
    \end{tabular}
\caption{The BLEU scores of adaptive $Y$ masking strategy.}
\label{tab:mask_y_relation}
\end{table}

\begin{table}[!]
\centering
\small
\begin{tabular}{l c | l c}
\toprule
\textbf{Masking Strategy} & \textbf{BLEU} & \textbf{Masking Strategy} & \textbf{BLEU} \\
\midrule
Adaptive (Ours) & \textbf{34.84} &Uniform  & 34.53 
 \\
+ same ratio & 34.65 &
Glancing  & 34.68 \\
+ 3 step & 34.50& 
Glat & 33.72 \\
+ exposure bias  & 34.79 &
Glat + fix-x (0.1) & 33.64\\
+ confidence-based  & 33.85 &
Glat + ada-x & 33.35\\
\bottomrule
\end{tabular}
\caption{Comparison of adaptive $Y$ masking with different constraints and related methods.}
\label{tab:adaptve Y}
\end{table}

\noindent{\textbf{The Effect of Adaptive $Y$ Masking.}}
To better understand the two-step training scheme and how to guide model training, we analyze the effect of different masking and training settings, and notice that we all keep the uniform masking strategy in the first step as the original CMLM.
First, we use uniform sampling to replace adaptive $\psi$ sampling in the second masking step.
Then we also keep the masking ratio $\alpha_{dec}=\beta$ to verify whether the masking ratio is critical for model training.
Besides, we use an adaptive strategy to train three steps which simulate the multi-step inference scenarios. We also test the impact of whether recover ground truth tokens or keep the predicted token in the second training step.
Since the masking tokens are chosen by prediction confidence during inference, we also apply confidence-based masking during training to further verify our adaptive $Y$ masking.
Moreover, we also compare our adaptive $Y$ masking with the glancing masking strategy proposed in GLAT to improve the one-pass decoding. 
The results are shown in Table~\ref{tab:adaptve Y}. 
We can observe that adaptive masking outperforms uniform masking in the second-step training, and the uniform masking seems to bring little improvements compared with adaptive $X$ masking (34.48). This also indicates that although AMOM may expand training expenses, adaptive $Y$ masking is truly valuable, and the performance improvements do not come from more updates. 
Moreover, results also reflect that two-step refinements are enough for model training without the necessity for more steps. 
Besides, using model prediction instead of ground truth can effectively reduce the problem of exposure bias, and introducing a confidence-based masking strategy does not bring improvements. 
Compared with GLAT, adopting the glancing masking as the second step masking strategy also performs better than uniform masking but is inferior to our adaptive $Y$ masking. 
Besides, if we directly adopt glancing masking and one-step training the same as GLAT (Glat), the performance declines, and further combining it with encoder masking even harms the performance.
This indicates that our methods play a different role compared with GLAT.

\begin{figure}[!htb]
\centering
\includegraphics[scale=0.085]{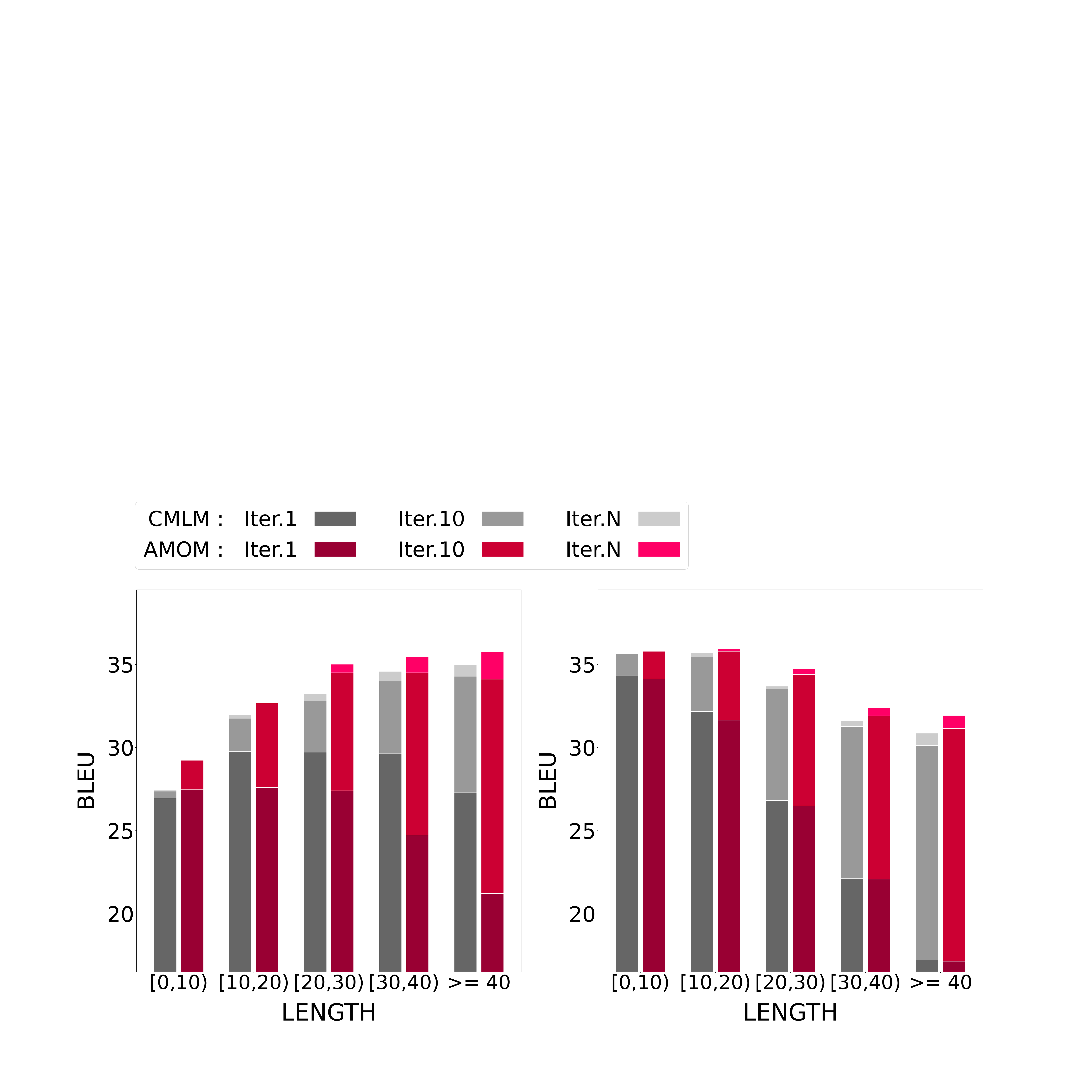}
\caption{Comparison between different source language sentence length and decoding iterations.}
\label{fig:diff_iter}
\end{figure}

\noindent{\textbf{More Iterations for Long Sequence.}}
\label{sec:more_iters}
For long source input sentences, it is almost impossible to obtain a fluent and relatively correct result for non-autoregressive machine translation models. 
It often requires multiple iterations to refine the translation results. Therefore, the ability to refine is a crucial evaluation criterion for a model. 
First, we compare the BLEU scores of AMOM and CMLM in different iterations steps, as shown in Appendix.
We can see that the AMOM outperforms the CMLM model when the iterations step increases, which proves that an adaptive masking strategy can enhance refinement ability.
In addition, we make a comparison of results with different source sentence length $N$ and different decoding iterations $\text{T}$ on two two datasets (IWSLT DE$\to$EN and WMT EN$\to$RO).
We split each dataset into five segments according to sentence length and run inference three times according to different steps $N \in [1, 10, N]$.
In Figure~\ref{fig:diff_iter}, we present the improvements of more decoding steps with different colours. Results show that AMOM exhibit significant gain than vanilla CMLM with more steps, e.g., although the performance of AMOM in Iter.1 is inferior than CMLM, it all outperforms CMLM in Iter.10, especially for long sentences. 
We can also find that long sentences often require more decoding steps, and AMOM perform better.

\section{Related Work}

\noindent{\bf Iterative-based Non-autoregressive Sequence Generation.}
Non-autoregressive models have attracted an increasing attention in recent years due to their efficient decoding, but the improvements in decoding speed come at the expense of generation quality.
Thus, iterative-based non-autoregressive (NAR) models~\cite{lee2018deterministic,gu2019levenshtein,saharia2020non,geng2021learning,lu2022insnet} are proposed to achieve a better trade-off between the inference speedup and generation quality.
\citeauthor{lee2018deterministic} first propose the iterative model which aims refine the noised target sequence. Later, insertion and deletion operations are introduced in each decoding iteration to create the final translation. Among these iterative NAR methods, the conditional masked language model (CMLM)~\cite{ghazvininejad2019mask} is widely-used owing to its promising performance when using the mask-predict strategy.
In particular, CMLM leverages the masked language model objective to guide model training and iteratively masks and predicts tokens during inference. 
Many recently works have achieved performance improvements based on CMLM~\cite{guo2020jointly,huang2022improving}.
Recently, \citeauthor{savinov2021step} proposed step-unrolled denoising
autoencoder which adopts denoising operation in each iteration.


\noindent{\bf Masked Language Model.}
The masked language model (MLM) first introduced by BERT~\cite{devlin2018bert} has become the essential component of various popular pre-training methods~\cite{song2019mass,liu2019roberta,dong2019unified,joshi2020spanbert,li2022universal,xu2021bert}. 
Its standard paradigm is to select some tokens in the source sequence by different strategies and then replace them with a \texttt{[mask]} token, and then the model is trained to predict the masked tokens. 
Since the masking strategy is significantly essential for these model, different masking strategies are served as different learning methods.
As BERT is served as a single Transformer encoder and a monolingual framework, there are limitations in various applications, such as machine translation. Then much progress has been made to extend the applications of masked language modeling strategy~\cite{guo2020incorporating,zhu2020incorporating,li2022universal}. 
The CMLM-based non-autoregressive models can also benefit from it by introducing a uniform masking strategy in training and a mask-predict decoding strategy during inference~\cite{ghazvininejad2019mask}. 
However, only few improvements on masking strategies are explored for CMLM.
In this work, we further design a simple yet effective adaptive masking over masking method on both the encoder and decoder sides to enhance the CMLM training for better refinement capability during inference.

\section{Conclusion}
In this paper, we present an adaptive masking over masking (AMOM) strategy to enhance the conditional masked language model (CMLM) for non-autoregressive sequence generation.
Our AMOM only contains two masking operations in model training without modifying the model structure or changing the inference schedule.
Extensive experiments on different sequence generation tasks indicate our proposed AMOM can yield significant performance improvement over the original CMLM model and even outperform the strong autoregressive (Transformer) counterpart on 7 NMT benchmark datasets and achieves SOTA performance on WMT16 EN$\to$RO, \textbf{34.82} BLEU on WMT16 RO$\to$EN, and \textbf{34.84} BLEU on IWSLT De$\to$En.
Due to the limitation of computational resources, we only test our AMOM for the CMLM model. 
In the near future, we will design more elegant AMOM strategies and explore their effectiveness on different NAR frameworks. 
We also will extend our AMOM to other types of masked language models, both in the pre-training and fine-tuning stages.

\section{Acknowledgments}
Ruiyang Xu contributes equally with Yisheng Xiao.
Juntao Li is the corresponding author.
This work is supported by the National Science Foundation of China (NSFC No. 62206194), the Natural Science Foundation of Jiangsu Province, China (No. BK20220488), and the Project Funded by the Priority Academic Program Development of Jiangsu Higher Education Institutions.
This work is also supported by Beijing Academy of Artificial Intelligence (BAAI).

\bigskip

\bibliography{aaai23}

\end{document}